\documentclass[preprint,review,12pt]{elsarticle}
\journal{ }
\usepackage[numbers]{natbib}
\usepackage{amssymb}
\usepackage{lineno} 
\usepackage[mathscr]{eucal} 
\usepackage{amsmath} 
\usepackage{algorithmic} 
\usepackage[linesnumbered,lined,boxed,commentsnumbered,ruled]{algorithm2e}
\usepackage{booktabs}
\usepackage{float}
\usepackage{subcaption}
\usepackage{comment}
\usepackage{multirow}
\usepackage[flushleft]{threeparttable}

\usepackage{xcolor}
\usepackage{color,soul}

\usepackage{csquotes}

\usepackage{setspace}

\RequirePackage[pdftex,
      colorlinks={true},
      linkcolor={Blue3},
      urlcolor={Blue3},
      citecolor={Blue3},
      pdfstartview={FitH},
      bookmarks={true},
            pdfauthor={Author},
            pdftitle={Title},
            pdfsubject={Subject}
            plainpages=false,
            pdfpagelabels
            ]{hyperref}

\usepackage[framemethod=tikz]{mdframed}
\mdfdefinestyle{mpdframe}{
    frametitlebackgroundcolor   =black!15,
    frametitlerule              =true,
    roundcorner                 =0pt,
    middlelinewidth             =1pt,
    innermargin                 =0.3cm,
    outermargin                 =0.3cm,
    innerleftmargin             =0.3cm,
    innerrightmargin            =0.3cm,
    innertopmargin              =0.5cm,
    innerbottommargin           =0.5cm
}

\newcommand{\htp}[1]{{\sethlcolor{yellow}\hl{#1}}}
\newcommand{\hfp}[1]{{\sethlcolor{cyan}\hl{#1}}}
\newcommand{\hfn}[1]{{\sethlcolor{magenta}\hl{#1}}}
\newcommand{\speaker}[1]{\textbf{#1}}

\MakeOuterQuote{"}

\begin{document}
\let\WriteBookmarks\relax
\def\floatpagepagefraction{1}
\def\textpagefraction{.001}

\title{Word-level Text Highlighting of Medical Texts for Telehealth Services}

\author{Ozan Ozyegen} 
\ead{oozyegen@ryerson.ca}
\author{Devika Kabe} 
\author{Mucahit Cevik}


\address{Data Science Lab at Ryerson University, Toronto, ON M5B 1G3, Canada}

    
    
    
    
    

\begin{abstract}
The medical domain is often subject to information overload. The digitization of healthcare, constant updates to online medical repositories, and increasing availability of biomedical datasets make it challenging to effectively analyze the data. This creates additional work for medical professionals who are heavily dependent on medical data to complete their research and consult their patients. 
This paper aims to show how different text highlighting techniques can capture relevant medical context. This would reduce the doctors' cognitive load and response time to patients by facilitating them in making faster decisions, thus improving the overall quality of online medical services. 
Three different word-level text highlighting methodologies are implemented and evaluated. The first method uses TF-IDF scores directly to highlight important parts of the text. The second method is a combination of TF-IDF scores and the application of Local Interpretable Model-Agnostic Explanations to classification models. The third method uses neural networks directly to make predictions on whether or not a word should be highlighted. The results of our experiments show that the neural network approach is successful in highlighting medically-relevant terms and its performance is improved as the size of the input segment increases. 
\end{abstract}


\maketitle

\nolinenumbers
\section{Introduction}



Since the start of the 20th century, people have been using telecommunication technologies to access health care services. This preference of communication over in-person visits has seen a gradual growth in popularity over the years. 
More recently, as a result of the COVID-19 pandemic, the demand for these services has significantly increased. 
This is especially the case with the current, ongoing lockdowns, as many people find themselves staying at home and using telehealth services to seek medical advice. 
The increasing demand for telehealth services presents opportunities for innovation in this area.

\textit{Information overload} in the health care sector is a growing problem.
Medical professionals and researchers rely on information from many resources such as clinical notes and scientific literature to accomplish their work. 
It is crucial for them to have quick and efficient access to up-to-date information. 
The information overload that these professionals face is mainly due to the enourmous amount of available unfiltered information. 
As a result, more time is spent searching for medical data and filtering out the ones that are not important.
This problem manifests itself in telehealth services as well. 
Medical professionals need to read a great deal of information about their patients who are seeking medical advice. 
Furthermore, when the doctors interact with multiple patients, they must keep track of each conversation. 

Automatic text highlighting of medically relevant information has potential to significantly improve the effectiveness and efficiency of healthcare practitioners since physicians would spend less time examining full text and only focus on key words and segments~\cite{dudko2018natural}. 
Many other domains have benefited from automated text highlighting.
\citet{ramirez2019understanding} has found that text highlighting can reduce the decision time to almost a half in crowdsourcing tasks. 
Researchers in information management and psychology have shown that text highlighting can improve the reading time \citep{wu2003improving}.
Previous studies have also examined the benefits of text highlighting in other areas such as supporting workers in the digitization of tasks by highlighting key fields \citep{alagarai2014cognitively}, requesting highlights as evidence and reasoning to support judgements \citep{schaekermann2018resolvable}, recommending text passages to facilitate the job of text annotators \citep{wilson2016crowdsourcing}, explaining the output to different machine learning models \citep{nguyen2018comparing}, and highlighting key terms in medical documents \citep{dudko2018natural}.

Biomedical informatics present certain challenges for natural language processing techniques. For instance, text categorization literature on electronic health records is limited. This is most likely due to the lack of train and test data that consists of high quality, labeled, full-length documents~\cite{argaw2007general}. Moreover, the contents of electronic health record systems are largely composed of clinical notes in the form of unstructured, unclassified texts~\cite{moen2016comparison}.
In order to address these challenges, large biomedical datasets and other resources have been made available to the researchers.
The available resources contain a wide range of texts from medical databases, such as clinical reports and full papers published in medical journals. 
With the increased amount of available data, researchers can now train large machine learning models for summarization in the medical domain~\cite{moen2016comparison}.

With the amount of data at the disposal of medical researchers and physicians, text highlighting and document summarization have become increasingly popular in the medical domain.
One can utilize this information to study diagnoses and common symptoms in patients without having to examine full documents and transcripts between patients and doctors. 
Knowledge repositories such as the Unified Medical Language System (UMLS)~\citep{bodenreider2004unified} contain numerous medical terms, and can be exploited in several ways by summarization engines. 
The Metathesaurus forms the base of the UMLS and comprises over 1 million biomedical concepts and 5 million concept names, all of which stem from over 100 incorporated controlled vocabularies and classification systems.

In this study, we mainly focus on text highlighting and extractive summarization for an online chat service connecting patients and doctors.
This service allows the patients to contact board certified medical doctors, and communicate their symptoms and/or any health concerns.
It is a novel way to provide access to quality healthcare without the need of having to go to a hospital, or walk-in clinic, similar to that of telehealth.
This also reduces the barriers that people face when trying to receive healthcare including taking time off work and incurring high medical costs.
Regarding the COVID-19 pandemic, increased access to the medical services without needing in-person visits might save important resources, and has potential to reduce the infections considerably.

In an online medical chat service, doctors receive online messages from patients experiencing health issues and give recommendations. 
Since the realm of medicine is so extensive, it would be beneficial for doctors to read the important specifics of the problem, rather than the full text. 
In addition, this ensures that the doctors do not miss useful pieces of information hidden inside a large text.
The primary goal of this paper is to develop a mechanism that takes the messages from the patients and highlights important words and short segments to facilitate doctors in their responses. 
This will make the overall job easier for the doctors, and speed up the time it takes to read and respond to patient queries. 

We investigate three different approaches for text highlighting of medical texts. 
The first is by using TF-IDF scores directly to find important parts of the text. 
The second is a combination of TF-IDF scores and Local Interpretable Model-Agnostic Explanations (LIME).
The latter is a method used to interpret models after they make their classifications \citep{molnar2019interpretable}, and the former shows how much a datapoint is indicative of the prediction. 
The final approach uses neural network models directly to decide whether each word should be highlighted.

The main contributions of this study can be summarized as follows:
\begin{itemize}
    \item To the best of our knowledge, this is the first study on word-level text highlighting of medical texts. 
    We establish a strong baseline and open-source one of our datasets to encourage further work in this area.
    
    \item We propose two novel word-level text highlighting methods for medical texts. 
    The methodologies we present show how to leverage three different sources of information: large medical term corpora, existing relevant metadata and manually labelled samples.
    
    \item We evaluate five text highlighting models trained using three different aproaches on two medical text datasets to provide a detailed numerical study that shows the effectiveness of the proposed methods. 
    This way, our study also provides evidence on the effectiveness of standard approaches for medical text highlighting.
\end{itemize}

The rest of this paper is organized as follows. 
In Section~\ref{sec:background}, we provide a brief literature review which overviews existing text highlighting methods. 
In Section~\ref{sec:dataset}, we list the different datasets and their specifications used in our analysis. 
In Section~\ref{sec:methodology}, explain the different text highlighting methodologies we employed.
We provide the results of our detailed numerical study in Section~\ref{sec:results}, which also includes performance comparisons for the text highlighting methodologies. 
Lastly, in Section~\ref{sec:conclusion}, we discuss the findings of our study, highlight its weaknesses and, provide future research directions.

\section{Background}\label{sec:background}

There are two different approaches to generating summaries from text: extractive and abstractive \citep{afantenos2005summarization}. 
The extractive summarization extracts sentences or small segments directly from the source text whereas the abstractive summarization may contain segments that are not present in the source text. 
The extractive approach can further be divided into two: sentence-level and word-level summarization. 
Most research in extractive summarization focuses on sentence-level summarization~\cite{mehta2016extractive}. 
The task of sentence selection can be considered as an information retrieval task \citep{reeve2006concept}, where all sentences within a text are evaluated through a scoring mechanism, and the highest scoring sentences are selected as being the most relevant to a summary. 

A common issue in extraction-based summarization is how to determine which sentences must be kept in a summary and which must be excluded \citep{moen2016comparison}. 
Disagreement among annotators can arise from various reasons including missing context, imprecise questions, contradictory evidence,
and multiple interpretations stemming from diverse levels of annotator expertise \citep{gurari2017crowdverge}. This is why it is important to measure the level of agreement to reach a consensus on what should be highlighted. 
For measuring inter-annotator agreement, Cohen's Kappa and Krippendorf's alpha are two popular measures~\cite{sanchan2017gold}. 
To achieve consensus among the physicians, \citet{van2007assessing} conducted a series of structured interviews with five Department of Medicine residents at New York-Presbyterian Hospital. The medical record of each physician is observed to identify sentences which are relevant for summarizing a patient’s history.
The physicians were provided with complete, de-identified patient records for three common general medical admissions. 
For each case, they were asked to acquaint themselves with the patient and then to underline information crucial to describing the patient to a colleague. 

Training word-level extractive summarization models requires a strategy to identify important segments of text in the dataset.
Term frequency-inverse document frequency (TF-IDF) is a widely used numerical statistic which can reflect how important a word or a segment of text is to a document in a corpus. 
\citet{laban2020summary} use TF-IDF to identify important keywords for training a reinforcement based abstractive summarization model.

\citet{moen2016comparison} propose and implement four novel automatic extractive summarization methods using electronic health records from patients in Finland. 
These methods were developed specifically with the clinical domain in mind.
The first approach is called repeated sentences. 
The underlying hypothesis for this method is that information that is repeated multiple times throughout an instance is the most important information to include in a summary. 
The second approach is called case-based, which retrieves existing cases with similar content. 
The underlying hypothesis is that patients with similar instances have similar content in their discharge summaries. 
The sentences from these discharge summaries are then treated as the central `topics' for what to include in the summary. 
The third approach is called translate, which aims to construct a type of translation system that can map sentences in clinical notes to the most probable sentences to be found in an accompanying discharge summary, based on translation statistics learnt from a large clinical corpus. 
The final approach is called composite, which consists of combining the above three methods, normalizing the scores and selecting the top sentences for the final summary.

Using domain concepts to form summaries is another method for sentence extraction. 
For instance, \citet{dudko2018natural} present a new approach of data retrieval from documents in the medical domain. 
This approach generates a summary of the document and highlights the most important words using document ontology which includes information about all recognized keywords and concepts in the document. 
Document ontology creation also includes the measure of relevance for each concept which is calculated through assigning points to the concepts \cite{dudko2019information}. 
The approach relies on the identification of medical keywords within the document text first, and then accumulating them into a document ontology in the form of a graph. 
This is used to produce the document summary. 
The keyword recognition in the text of a given document is performed using natural language processing techniques. 
Every identified keyword receives one point and if a keyword is repeated multiple times, every occurrence provides an additional point. 
The voting approach is then used to highlight terms with the highest number of points. This proposed approach does not try to suggest a diagnosis or make any other kind of decision automatically. It demonstrates which concepts are written about in the document and which keywords explain these concepts, whereas our methods form a prediction on the text, and then highlight based on the given prediction.
Similarly, \citet{reeve2006concept} proposed the frequency of domain concepts as a method to identify important sentences within a full-text to summarize biomedical texts. 
This method scores sentences based on a set of features, and extracts sentences with the highest scores to form a summary.


Extractive summarization is also possible using deep learning approaches.
\citet{verma2017extractive} propose a text summarization approach for factual reports using a deep learning model, that consists of three phases, namely feature extraction, feature enhancement, and summary generation. 
They use deep learning in the second phase to build complex features out of simpler features extracted in the first phase. Specifically, they employ a Restricted Boltzmann Machine to enhance those features to improve the resulting accuracy without losing any important information.
They score the sentences based on those enhanced features, and construct an extractive summary accordingly.


Aside from standard extractive summarization methods, another method to perform text highlighting is through LIME. 
It is a model-agnostic approach and involves training an interpretable model on samples created around a specific data point by perturbing the data \citep{ribeiro2016model}. 
Once a prediction is made, LIME outputs the features that are indicative of the prediction. 
\citet{nguyen2018comparing} uses LIME to evaluate the interpretability of text classification models. 
\citet{ribeiro2016should} emphasize the importance of being able to trust a model's predictions. 
They employ LIME to evaluate text classification models by using human subjects to measure the impact of explanations on trust and associated tasks.
\citet{moradi2020explaining} use LIME in the context of medical text classification, where they produce explanations for predictions on a disease-treatment related information extraction task.

In our analysis, we employ LIME as a method of highlighting important words relevant to patients' conditions in their messages to doctors. 
To highlight the important words via LIME, patients' messages are first classified into a category, and then LIME is utilized to extract words relating to the category. 

\section{Datasets}\label{sec:dataset}
We use two different datasets in our analysis. 
The first one is the propriety medical chat dataset that consists of chat messages between patients and doctors, while the second one is the open source MTSamples data which includes medical transcriptions for the patients treated in a hospital setting.

\subsection{Medical Chat Dataset}
This dataset is obtained from a telehealth service company. 
It contains written conversations between patients and doctors, where patients explain how they feel and symptoms they have, according to which doctors provide medical advice. 
In most cases, the doctor is able to provide recommendations without physical examination.
However, when further tests are needed to make a proper diagnosis, the patient is advised to go to a hospital or walk-in clinic.
The dataset includes a collection of 33,699 doctor-patient conversations between October 6, 2019 and July 15, 2020, and 901,939 messages exchanged between them. 

A conversation always starts with a patient message automatically generated using a predefined questionnaire that the patient fills out. 
Based on this form, an important metadata called \textit{issue category} is created. 
This property describes the type of the issue patient has, such as pregnancy or COVID-19, and every conversation is tagged with an issue category.
The medical chat dataset is used to derive two datasets: \textit{Medical Chat Classification Dataset} and \textit{Medical Chat Highlighting Dataset} to allow experimentation with different methodologies.

\subsubsection{Medical Chat Classification Dataset}
This dataset is designed for a multi-class classification task. 
The goal is to predict the issue category from a doctor-patient conversation.
There exists eight issue categories in the dataset. These are cold or flu, COVID-19, gastrointestinal, pregnancy, sexual health, skin, pain management and other. 
In our study, the last two categories were omitted, because the pain management category has very few samples and the other category contains many mislabeled samples. 

It is important to note that we use this dataset in order to leverage the existing labels (issue categories) available in the dataset. 
The methodology proposed using this dataset allows us to design a text highlighting model without any additional annotations for text highlighting.

\subsubsection{Medical Chat Highlighting Dataset}
This dataset contains a manually labelled subset of the medical chat dataset for the text highlighting task. 
A subset of 350 conversations were randomly selected from the last two months of the available data. 
The data contains around 8,330 messages since each chat includes on average 23.8 messages. 
The first 300 conversations and the last 50 were separated as the training and test sets, respectively. 

In the text highlighting task, each word has a binary label, 1 for highlighted and 0 for unhighlighted. 
To minimize the differences among annotations, the doctors prepared guidelines for the annotation process. 
For instance, it was required to highlight numbers, body parts and the gender of the patient. 
The annotation of the training set was split among three annotators where each annotator only annotated a subset of the training set. 
The test set samples were annotated by all three annotators to measure the inter-annotater agreement. 
The agreement level between the annotations is high in terms of Krippendorf's alpha $\alpha = 0.82$, demonstrating the consistency of the proposed annotations.

\subsection{MTSamples Highlighting Dataset}
The second dataset we use is derived from an open-source dataset called MTSamples~\cite{yetisgen2016automatic}. 
It is a large collection of publicly available transcribed medical reports. 
It contains sample transcription reports, provided by various transcriptionists for many specialties and different work types. 

The original dataset has over 2300 transcriptions. 
However, most of the samples consist almost entirely of medical terms. 
Thus, they are not usable for evaluating a medical text highlighting model. 
For this reason, we picked 100 transcriptions from this dataset that are of the similar nature to the Medical Chat Highlighting dataset. 
The following quote is an example of a medical transcription from the MTSamples dataset: 
\begin{displayquote}
\textit{``Patient having severe sinusitis about two to three months ago with facial discomfort, nasal congestion, eye pain, and postnasal drip symptoms.''}
\end{displayquote}

It is important to note that this dataset was used only for evaluating the text highlighting models, and it was not used for training. 
The models trained on the previously mentioned datasets were also tested on this dataset to evaluate the usability of the models on different medical texts as shown in Figure~\ref{fig:methods}.
Similar to the medical chat highlighting test set, each sample in this dataset is labelled by three annotators to measure the annotator agreement. 
The inter-annotator agreement level is again high in terms of Krippendorf's alpha $\alpha = 0.85$.
Finally, we open source this dataset to encourage further work in text highlighting of medical texts~\footnote{Github link will be provided here in the final version of the paper}.

\subsection{Data Preprocessing}
We applied the following preprocessing steps to improve the quality of the inputs provided to the models. 
\begin{itemize}
    \item OpenNMT tokenizer~\cite{klein-etal-2017-opennmt} is used for tokenization. 
    This tokenizer supports reversible tokenization by annotating tokens or injecting modifier characters. 
    We further implemented a wrapper tokenizer around the OpenNMT tokenizer to easily map the predicted highlights back to the original raw texts.
    
    \item We decided to keep stopwords in the datasets since these words are sometimes part of the highlights. 
    We observed that the model performances drop if we remove the stopwords, since the model is unable to highlight stopwords such as ``of'' or ``my''.
    
    \item Other preprocessing steps include removing URLs, converting all characters to lowercase and finally, removing non-Unicode characters and line breaks.
\end{itemize}


\section{Methodology}\label{sec:methodology}
\begin{figure}[!ht]
    \centering
    \includegraphics[width=1\textwidth]{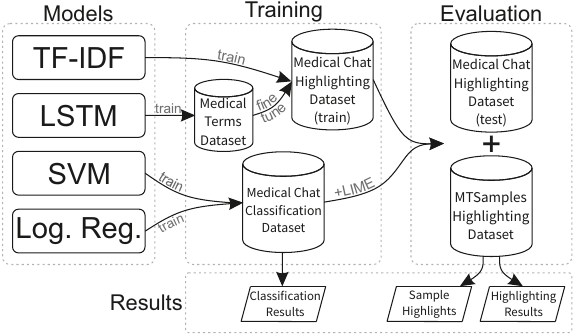}
    \caption{Experimental setup, showing how the models are trained and evaluated.}
    \label{fig:methods}
\end{figure}

In this section we describe our experimental setup which is summarized in Figure~\ref{fig:methods}. 
We train five text highlighting models using three different methodologies.
The first one is the baseline model that uses TF-IDF scores to find the important segments in the text. 
Unlike the other methodologies, TF-IDF only uses the Medical Chat Highlighting dataset for training. 
In the LIME approach, we leverage the existing labels on the dataset for the text highlighting purpose. 
Here, we first train our models on a classification task. 
Then, in the evaluation stage, we use the feature importances produced by the LIME method to decide which segments should be highlighted. 
Finally, in the neural network based approach, we first train the models on a large medical terms corpus, and then \textit{fine tune} the models on the Medical Chat Highlighting dataset.

\subsection{TF-IDF}
TF-IDF works by determining the relative frequency of words in a given document compared to the inverse proportion of that word over the entire document corpus \citep{ramos2003using}. 
This calculation then determines the relevance of a given word to a specific document. 
Therefore, words that are common in a single or a small group of documents tend to have higher TF-IDF scores compared to common words such as "i", "me", "to", and "the". 
There are varying procedures for implementing TF-IDF.
We summarize the overall approach as follows. Given a document collection $D$, an n-gram $t$ and an individual document $d$ $\in$ $D$, the TF-IDF score of each $n$-gram in a given document is calculated as follows:
\begin{equation}
    \label{TFIDF}
    tfidf(t,d,D) = f_{t,d}*\text{log}(|D|/f_{t,D})
\end{equation}
where $f_{t,d}$ is the frequency of $n$-gram $t$ in document $d$ and $f_{t,D}$ is the number of documents that contains the $n$-gram $t$, also referred as inverse document frequency.
For training the TF-IDF model, we use all the unigrams, bigrams and trigrams in the medical chat highlighting dataset. 
After training the model, we evaluate the model by computing precision, recall, ROC-AUC and PR-AUC scores. 
For precision and recall calculation, we use a fixed threshold of 0.01 for both datasets. 
While precision and recall shows the model performance at a specific threshold, ROC-AUC and PR-AUC measures the overall model performance.
For training the TF-IDF model on medical chats, we can either treat each conversation or each message as a document. 
We found that using each conversation as a document tends to perform better mainly because the individual messages can be very short. 
It is important to note that the dataset size can also be an important factor in training a TF-IDF model. 
Thus, we first trained the TF-IDF model on larger subsets of the Medical Chat dataset which did not lead to any improvement over using the training set of the Medical Chat Highlighting dataset.

\subsection{LIME}\label{lime-section}
An important disadvantage of the TF-IDF method is that no contextual information is provided to the model. 
However, in some cases, the dataset may contain labels that can be leveraged to solve the text highlighting task. 
In the LIME approach, we present a strategy for using the existing labels in a dataset for text highlighting.

LIME~\cite{ribeiro2016should} is a popular post-hoc local interpretability method. 
The algorithm takes a pretrained model, a multi-class classification model in our case, and a sample from the dataset. 
It then trains a linear classifier that approximates the models' predictions locally. 
Afterwards, the coefficients of the linear classifier are used to identify the importance of the features in the given sample.

The process for LIME can be seen in Figure~\ref{fig:methods}. 
First, a multi-class classification model is trained on the Medical Chat Classification dataset. 
We experimented with three algorithms: Logistic Regression (LR), Support Vector Machine (SVM) and Long Short Term Memory Networks (LSTM). 
For LR and SVM, vectorized TF-IDF scores of unigrams, bigrams and trigrams are provided to the model. 
For the LSTM model, we use the Word2Vec representation of each word. 
After training the models, to find the highlights for a sample message, the LIME explainer is used to explain the predictions of the classifiers. 

The LSTM results were omitted from our analysis with LIME, because, for LSTM models, we observed that the feature importances tend to be distributed among more words. 
A similar observation has been made in \cite{nguyen2018comparing}. 
The more distributed representation of feature importances results in a significantly lower text highlighting performance using the LIME method. 
LR and SVM models on the other hand tend to assign much higher importance to a few words that are important for the prediction.

\subsection{LSTM}
The Long Short Term Memory (LSTM) network~\cite{hochreiter1997long} is a very popular choice for sequential data processing because of its ability to capture long and short term dependencies. 
\citet{graves2005framewise} found that LSTM's effectiveness can be improved by splitting the LSTM cells where one part is responsible for the positive time direction, and the other part is responsible for the negative time direction also known as Bidirectional LSTMs.

For the text highlighting task, we propose a Bidirectional LSTM model with three layers. 
The first layer is an embedding layer that converts the words into semantically meaningful vector representations. 
In order to capture the semantics of the medical terms, we use pretrained Word2Vec embeddings trained on PubMed abstracts and PubMed Central full text documents~\cite{moen2013distributional}. 
The second layer of the network is the Bidirectional LSTM layer that processes the vector representations of the input words. 
This layer has 256 LSTM cells, a sigmoid activation and a dropout of 0.2 is applied for regularization. 
Finally, a linear layer with sigmoid activation is applied to each temporal slice of the LSTM output to predict the highlights. 
Binary cross-entropy loss and Adam~\cite{kingma2014adam} optimizer were used for training the network.

We also experimented with different hyperparameter values such as the number of LSTM layers, the number of LSTM cells, activation functions, and dropout values, which did not lead to a significant increase in the prediction performance.
In addition, we explored the importance of the context for the text highlighting task. 
For this reason, the LSTM architecture is trained in two different approaches.
In the unigram version, only one word at a time is provided to the model. 
The model takes a single word and tries to predict whether it will be highlighted or not. 
On the other hand, in the $n$-gram version, the full message is provided to the model. 
This allows the network to consider contextual information as it makes predictions for each word in the message.

\paragraph*{Fine Tuning}
To improve the performance of the LSTM models, we applied a fine tuning strategy. 
Specifically, the models are first trained on the \textit{Medical Terms dataset} and later fine tuned on the Medical Chat Highlighting dataset.
The Medical Terms dataset contains medical and non-medical terms. 
For medical terms, two data sources were used, the Unified Medical Language System (UMLS)~\citep{bodenreider2004unified} and the International Classification of Diseases (ICD) codes \citep{afzal2018natural} which have 661,675 and 12,704 unique medical terms, respectively. 
The non-medical terms datasets consist of a variety of texts from electronic books \citep{beard2020history, doyle1892adventures, tolstoy2016war} and contain 25,553 unique words. 
The models trained on the Medical Terms dataset are then fine tuned on the Medical Chat Highlighting dataset. 
We also performed additional experiments to see the impact of Medical Chat Highlighting dataset size (the number of annotated training samples) on the text highlighting performance. 


\section{Results}\label{sec:results}
In this section, we summarize the results of our numerical study. 
We first compare the performances of five different methods used for medical text highlighting considering Medical Chat and MTSamples datasets.
Then, we provide text highlighting visualizations with the best performing method to show the practical use case of our approaches.
Lastly, we provide our observations on how the performance of the fine-tuned LSTM model changes with increasing number of annotated instances.

We performed all experiments using a workstation with i9-9990K 3.6GHz CPU, RTX2070 SUPER 8Gb GPU, and 128Gb of RAM, with Debian Linux OS. 
The Scikit-learn library is used for implementing TF-IDF, LR and SVM. 
For the neural network architecture implementations, Tensorflow 2 library is used.

\subsection{Medical Text Highlighting Results}
In this section, we first briefly define the two sets of performance metrics used to evaluate different algorithms. 
Then, we report the performance of the five models on the Medical Chat Highlighting and MTSamples datasets. 

The first set consists of precision and recall, which describes the "accuracy" of the model for a specific threshold (i.e., ``single threshold''). 
These metrics provide an intuitive sense on how well the models perform.
Precision indicates what percentage of the predicted word highlights are true highlights, and recall shows what percentage of the true highlights were succesfully found. 
The same thresholds can have very different precision and recall values on different models. 
Thus, we handpicked a reasonable threshold for each model. 
If two models had a similar performance on the same dataset, we chose a threshold where either the precision or recall values are similar to make it easy to compare models' performances.

The second set of metrics evaluate the performance of the models in a more general way (i.e., ``multi threshold''). 
AUC-ROC and PR-AUC scores are computed by measuring the performance of a classification model at various threshold settings. 
PR-AUC is preferred over AUC-ROC when there is severe class imbalance between the classes or when the minority class is more important~\cite{saito2015precision}. 
For text highlighting, one can argue that the positive (minority) class is more important, since these correspond to medical terms. 
Thus, we computed both AUC-ROC and PR-AUC scores which resulted in the same ranking between the models.

\begin{figure}[!ht]
    \centering
    \subfloat[LR]{\includegraphics[width=0.45\textwidth]{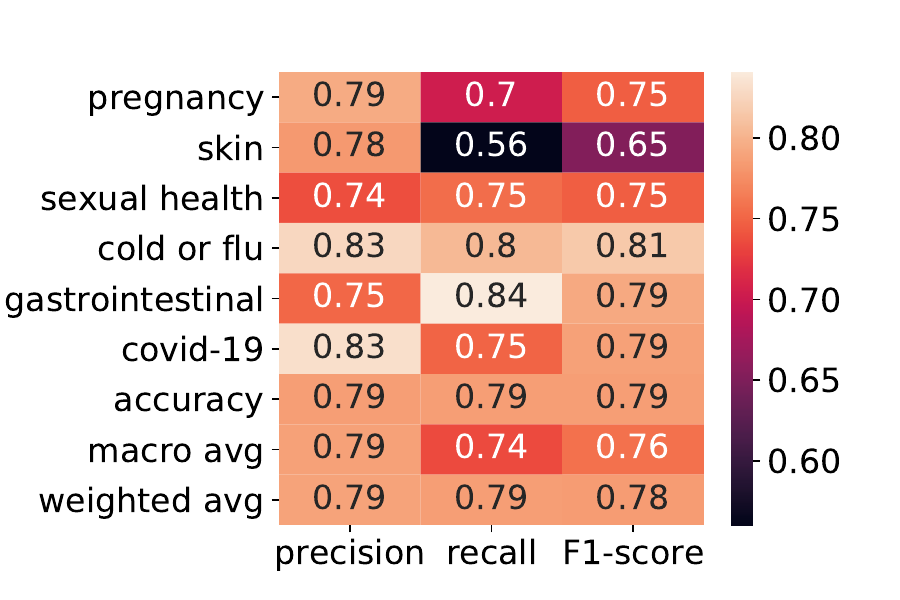}
    }
    \hfill
    \subfloat[SVM]{\includegraphics[width=0.45\textwidth]{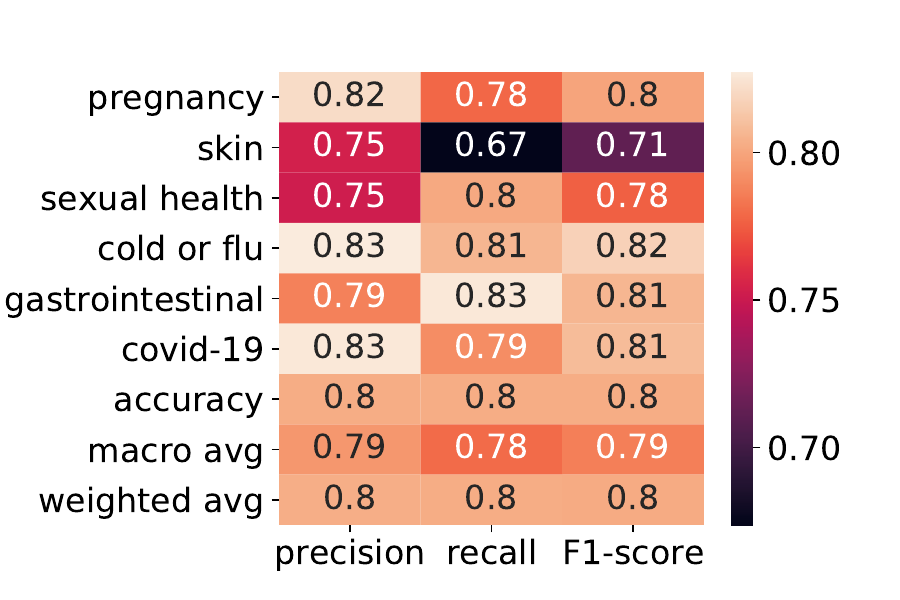}
    }
    \caption{Classification performance of the LR and SVM models on Medical Chat Classification dataset}
    \label{classification_results}
\end{figure}

As described in Section~\ref{lime-section},
LR and SVM models that are used for the LIME method are first trained on Medical Chat Classification dataset. 
The classification performance of these models are shown in Figure~\ref{classification_results}. 
SVM model outperformed the LR model for all the categories in terms of the F1-score. 

After all five models are trained on the corresponding datasets,
they are evaluated on the Medical Chat Highlighting and MTSamples datasets. For the Medical Chat Highlighting dataset, only patient messages were considered for evaluating the performance of the models because the aim of the proposed model is to automatically highlight patient messages, not doctor messages.
The results for the medical chat highlighting is shown in Table~\ref{tab:medical-text-table}. 
The LSTM models trained on the annotated samples outperformed the others, and the $N$-gram LSTM model performed slightly better than the unigram version of the model. 
The TF-IDF and LIME based methods performed significantly worse than the LSTM-based models. 
During manual evaluation of the sample highlights, we observed that both TF-IDF and LIME methods fail to highlight non-medical terms such as dates and numbers, which are considered to be important by doctors. 
This result is expected since dates and numbers were not considered as important features by both the LIME and TF-IDF methods.

\setlength{\tabcolsep}{7.5pt} 
\renewcommand{\arraystretch}{1.2} 
\begin{table}[!ht]
    \centering
    \scalebox{0.8}{
        \begin{tabular}{llcccccc}
\midrule
 & &
\multicolumn{2}{c}{\textbf{Single Threshold}} & &
\multicolumn{2}{c}{\textbf{Multi Threshold}}\\
\cline{3-4}\cline{6-7}\\[-0.9em]
\textbf{Model} & \textbf{Threshold} & \textbf{Precision} & \textbf{Recall} & & \textbf{ROC-AUC} & \textbf{PR-AUC} \\
\midrule
\multirow{1}{*}{TF-IDF} & 0.01 & 0.30 & \textbf{0.84} & & 0.79 & 0.40 \\
\multirow{1}{*}{SVM + LIME} & 0.005 & 0.29 & 0.53 & & 0.67 & 0.29 \\
\multirow{1}{*}{LR + LIME} & 0.005 & 0.31 & 0.61 & & 0.70 & 0.36 \\
\multirow{1}{*}{Unigram LSTM} & 0.5 & 0.72 & 0.82 & & 0.96 & 0.86 \\
\multirow{1}{*}{\textbf{N-gram LSTM}} & 0.5 & \textbf{0.81} & \textbf{0.84} & & \textbf{0.98} & \textbf{0.92} \\
\bottomrule
\end{tabular}
    }
    \caption{Evaluation of the models on Medical Chat Highlighting dataset. Bold numbers correspond to the best performing value for each metric.}
    \label{tab:medical-text-table}
\end{table}

For the MTSamples dataset on the other hand, almost all the highlighted terms are medically relevant terms, which provides a more fair comparison between the methods. 
The results for the MTSamples dataset is shown in Table~\ref{tab:MTSamples-results}. 
Again the LSTM models outperformed the alternative approaches that do not require any annotated samples. 
However, for the MTSamples dataset, the LIME based approaches outperformed the TF-IDF baseline. 
This indicates that the LIME based models are better able to capture medically relevant terms in the dataset than the TF-IDF model.

\setlength{\tabcolsep}{7.5pt} 
\renewcommand{\arraystretch}{1.2} 
\begin{table}[!ht]
    \centering
    \scalebox{0.8}{
        \begin{tabular}{llcccccc}
\midrule
 & &
\multicolumn{2}{c}{\textbf{Single Threshold}} & &
\multicolumn{2}{c}{\textbf{Multi Threshold}}\\
\cline{3-4}\cline{6-7}\\[-0.9em]
\textbf{Model} & \textbf{Threshold} & \textbf{Precision} & \textbf{Recall} & & \textbf{ROC-AUC} & \textbf{PR-AUC} \\
\midrule
\multirow{1}{*}{TF-IDF} & 0.01 & 0.80 & 0.05 & & 0.52 & 0.33 \\
\multirow{1}{*}{SVM + LIME} & 0.005 & 0.61 & 0.51 & & 0.69 & 0.53 \\
\multirow{1}{*}{LR + LIME} & 0.005 & 0.59 & 0.58 & & 0.72 & 0.58 \\
\multirow{1}{*}{Unigram LSTM} & 0.5 & \textbf{0.87} & 0.70 & & 0.95 & 0.91 \\
\multirow{1}{*}{\textbf{N-gram LSTM}} & 0.5 & \textbf{0.87} & \textbf{0.81} & & \textbf{0.96} & \textbf{0.92 }\\
\bottomrule
\end{tabular}
    }
    \caption{Evaluation of models on the MTSamples dataset. The models from the Table~\ref{tab:medical-text-table} are used without additional training. Bold numbers correspond to the best performing value for each metric.}
    \label{tab:MTSamples-results}
\end{table}

\begin{figure}[!ht]
    \centering
    \subfloat[ROC Curve]{\includegraphics[width=0.45\textwidth]{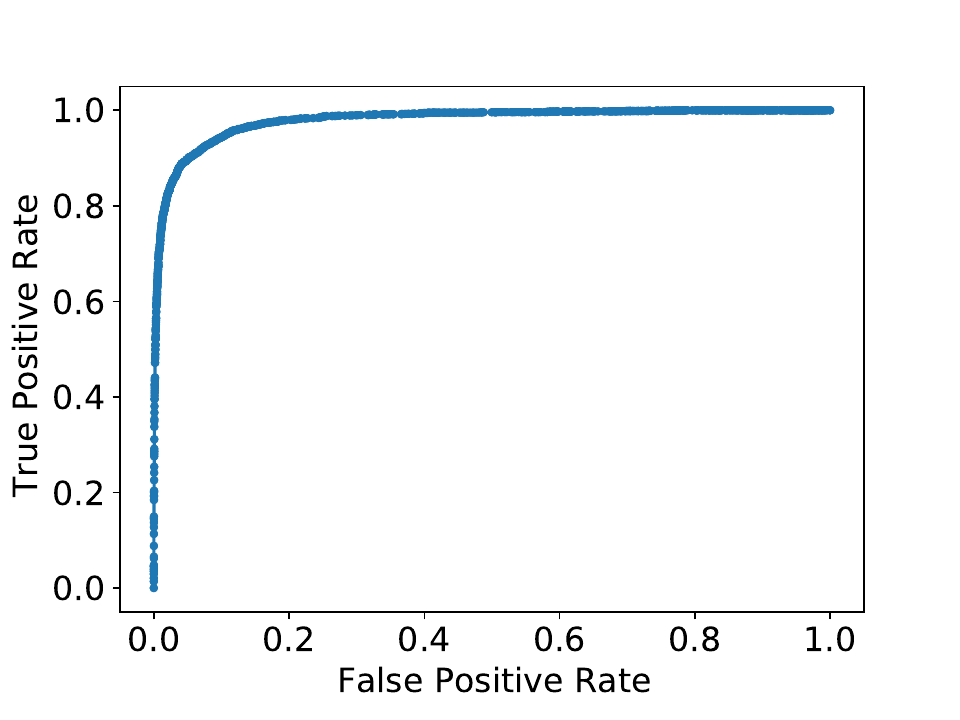}
    }
    \hfill
    \subfloat[Precision-Recall Curve]{\includegraphics[width=0.45\textwidth]{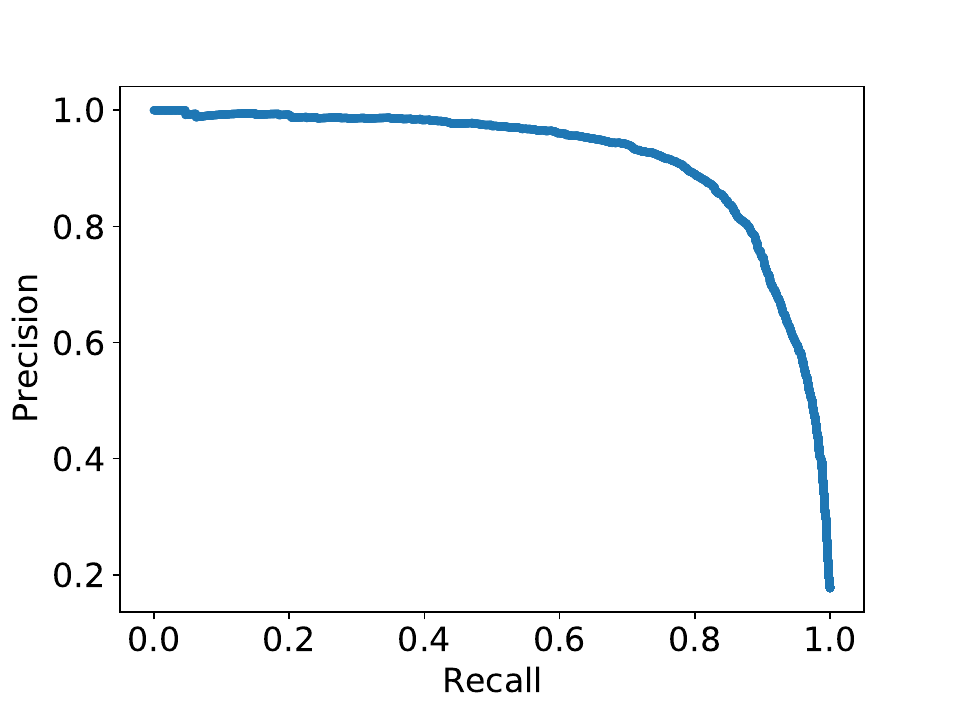}
    }
    \caption{Classification performance of the N-gram LSTM model on the Medical Chat highlighting dataset.}
    \label{fig:yd_curves}
\end{figure}

\begin{figure}[!ht]
    \centering
    \subfloat[ROC Curve]{\includegraphics[width=0.45\textwidth]{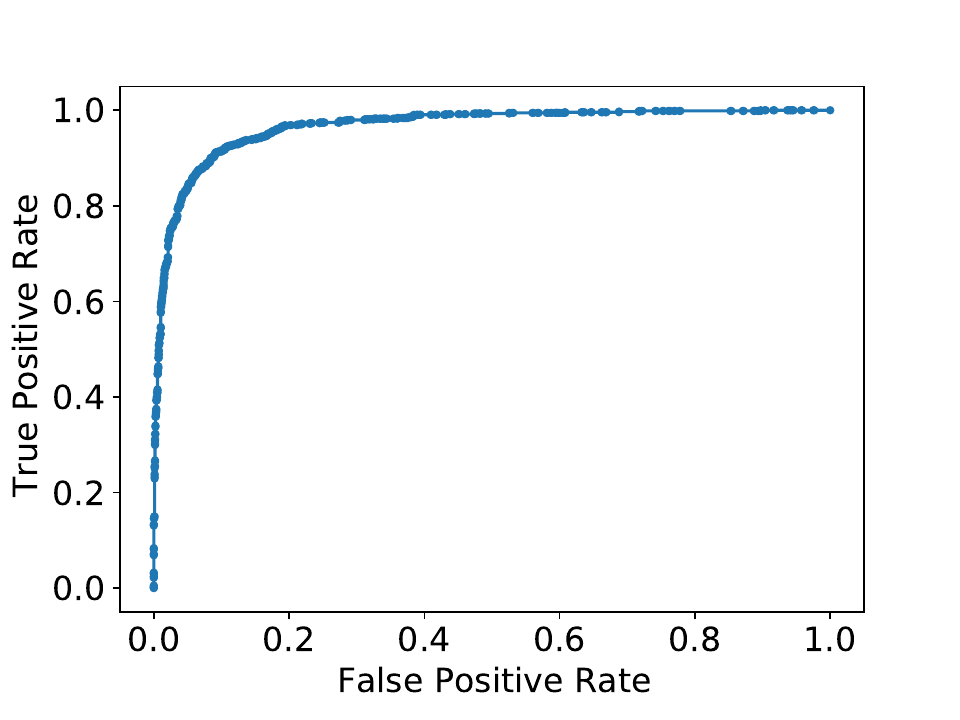}
    }
    \hfill
    \subfloat[Precision-Recall Curve]{\includegraphics[width=0.45\textwidth]{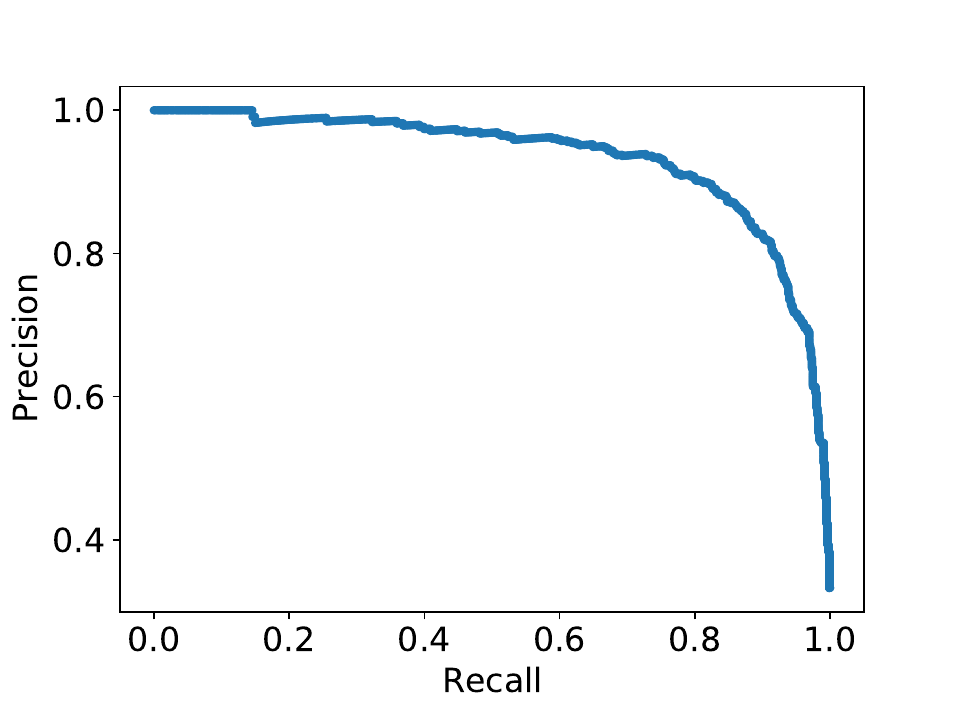}
    }
    \caption{Classification performance of the N-gram LSTM model on the MTSamples dataset.}
    \label{fig:mt_curves}
\end{figure}

Figures~\ref{fig:yd_curves} and \ref{fig:mt_curves} show the ROC and Precision-Recall curves for the n-gram LSTM model for the Medical Chat and MTSamples datasets. 
We note that the model performs similarly well for both sets.

\subsection{Visualizing the predictions}
To compare the overall performance of the models, ROC-AUC and PR-AUC metrics are used. 
After choosing the best model based on these metrics, we performed further evaluation of the model performance by visualizing the model predictions. Figures~\ref{chat-ex1} and Figure~\ref{chat-ex2} illustrate true positives (yellow), false positives (blue) and false negatives (magenta) on two random conversations from the Medical Chat Highlighting test set. 
The $n$-gram LSTM model is used with the default decision threshold of 0.5 for the visualizations. 
For privacy reasons, some of the names, numbers and drug names are obscured in the figures. 
From both samples, we can see that the model correctly identifies gender information, age, symptoms and drug information. 
In Figure~\ref{chat-ex2}, we see that the model is unable to capture some of the date related information and adjectives. 
We observed a similar misclassifications across all the models. 
Further analysis of the misclassified terms showed that there is some mislabelling of these term in the training data where phrases like "last week" are highlighted in some cases but, not highlighted in others. 
As such, we note that further verification of the training set can improve the performance of the models.

\begin{figure}[!ht]
\begin{mdframed}[style=mpdframe,frametitle=Random chat Id: 872577]
\speaker{Patient}: Hello Doctor! My name is [[NAME]]. Gender is \htp{Female}. Age is \htp{[[AGE]]}. My main issue is: \hfp{Terrible} \htp{stomach pain}, \htp{irritability}, \htp{headache}. \htp{Anemic} and taking \htp{ferrous sulface}. My current medications are: none                  \\
\speaker{Doctor}: Do you have \htp{constipation}? \\
\speaker{Patient}: Yes, I had a \htp{blood test} done to my \htp{extreme fatigue}. I have \htp{extremely low iron} \\
\speaker{Patient}: I am on \htp{ferrous sulfate} to raise my \htp{iron}. It makes me feel like I have the \htp{flu} \\
\speaker{Doctor}: Usually people who take the \htp{ferrous sulfate} tend to get \htp{constipated} \\
\speaker{Doctor}: Drink more water. Eat more fiber and exercise more. Take \htp{[[DRUG NAME]]}, an \htp{herbal laxative}. Take \htp{probiotic supplements}. \\
Conversation ended
\end{mdframed}
\caption{A random chat from the dataset where true positives are highlighted in yellow and false positives in blue.}
\label{chat-ex1}
\end{figure}

\begin{figure}[!ht]
\begin{mdframed}[style=mpdframe,frametitle=Random chat Id: 872492]
\speaker{Patient}: Hello Doctor! My name is [[NAME]]. Gender is \htp{Female}. Age is \htp{[[AGE]]}. My main issue is: Labs. My current medications are: none \\
\speaker{Patient}: Hello. I have a question about my \htp{bloodwork}. Especially in regards to \htp{coronavirus}. Do you know what the \htp{bloodwork} looks like for a person with \htp{COVID}? \\
\speaker{Doctor}: Yes. How many days post \htp{COVID 19}? \\
\speaker{Patient}: It has been \hfp{3 months}. I am concerned because I think I am still suffering from the \htp{virus}. I always \hfn{feel} \htp{sick} \\
\speaker{Doctor}: When did you last got the test done? It must have showed its \hfp{negative}? \\
\speaker{Patient}: Yes but I don't believe in those \htp{nasal tests}. I just had it done \hfp{last week}. So in regards to \htp{COVID patients bloodwork} - what values would be off? If any? Mine have a WBC of \htp{[[NUMBER]]}, mildly \hfp{elevated} \htp{neutrophils} (\htp{[[NUMBER]]}) mildly \hfp{low} \htp{lymphocytes} (\htp{[[NUMBER]]}) and \htp{platelets} of \htp{[[NUMBER]]}. My \htp{hemoglobin} was \htp{[[NUMBER]]} and that went down from \htp{[[NUMBER]]}.
\\
\speaker{Doctor}: Your \htp{blood work} is normal. You should not worry about the \htp{virus} anymore. If you had the \htp{virus} the symptoms may have worsen by now.
\\
\speaker{Patient}: How do you know? It can go into remission and then restart again can't it? All \htp{viruses} do.  \\
\speaker{Doctor}: You have to be \hfp{reinfected} in case of \htp{COVID 19}, it does not go in remission. Some \htp{viruses} live dormant, not this \htp{virus}. \\
\speaker{Patient}: I assumed it can be. Thanks for the clarification. \\
Conversation ended
\end{mdframed}
\caption{A random chat from the dataset where true positives are highlighted in yellow, false positives in blue and false negatives in red.}
\label{chat-ex2}
\end{figure}

\subsection{Fine-tuning LSTM models}\label{fine-tune-sec}
\begin{figure}[!ht]
    \centering
    \includegraphics[width=0.75\textwidth]{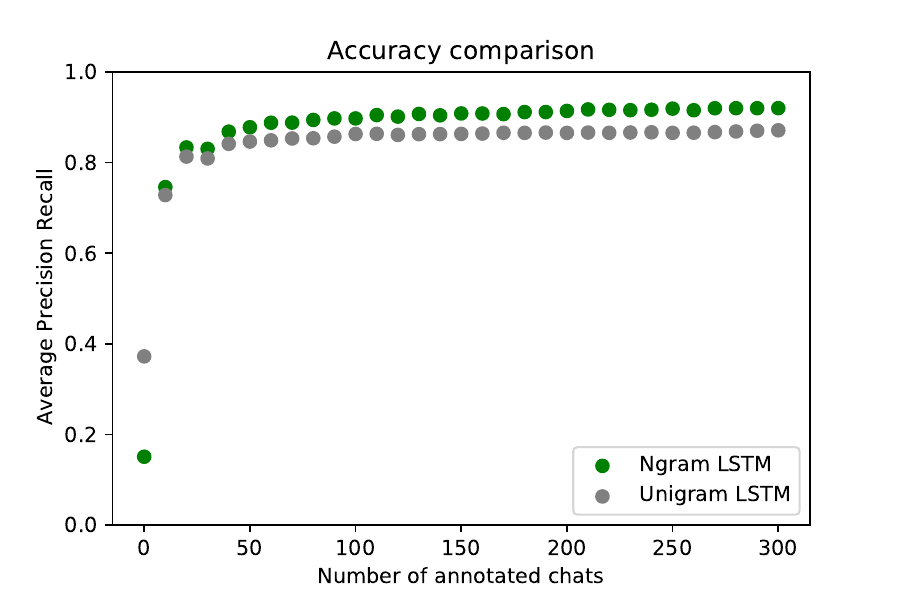}
    \caption{Increase in the accuracy (PR-AUC) as the LSTM models were fine tuned on the Medical Chat Highlighting dataset with more annotated data.}
    \label{fig:fine-tuning}
\end{figure}

In our analysis, we first trained the unigram and $N$-gram LSTM models on the Medical Terms dataset where the goal is to classify between medical and non-medical terms. 
This training step enabled the model to capture previously unseen medical terms. 
Then, the models are fine-tuned and evaluated on the Medical Chat Highlighting dataset. 
An important question is how the model performance changes based on the amount of labelled traning data.
To answer that question, we gradually increase the number of annotated chats on the Medical Chat Highlighting training data, and report the PR-AUC on the Medical Chat Highlighting test dataset. 
Figure~\ref{fig:fine-tuning} shows how the accuracy (PR-AUC) increases as the models were fine-tuned on a larger training set. 
For each dot in the figure, we first restore the weights of the model that is only trained on the Medical Terms dataset. 
Then, we train the model with a randomly selected subset of the Medical Chat Highlighting training dataset. 
The number of annotated chats provided for training increases in intervals of ten chats; starting from zero to three hundred chats. 

The results show the $N$-gram LSTM model outperforms the Unigram LSTM model in all cases except for the initial case where the models are only trained on the Medical Terms dataset. 
Since the Medical Terms dataset only contains medical and non-medical terms, not the full chat messages, the models perform significantly worse without fine tuning on the Medical Chat Highlighting dataset.

\section{Conclusion and Future Work}\label{sec:conclusion}
Telehealth companies are experiencing a rapid expansion as a result of the digitization of healthcare. 
The current demand for telehealth services present opportunities for innovation in this area. 
The automatic highlighting of medically relevant text can significantly improve the effectiveness and efficiency of healthcare practicioners. 
To the best of our knowledge, the word-level text highlighting of medical texts are explored for the first time in this study. 
By establishing a baseline and exploring various strategies, we investigate the feasibility of word-level text highlighting of medical texts in the telehealth domain. 
We propose two new word-level text highlighting methodologies for medical texts. 
The methodologies that we use leverage three different sources of information: large medical term corpora, existing relevant metadata and manually labelled samples. 
Furthermore, by evaluating five different text highlighting models on two medical text datasets, we provide a detailed numerical study that shows the effectiveness of the proposed methodologies.
Our results show that our best performing model, $n$-gram LSTM can successfully highlight the important information in medical texts, as evidenced by high precision and recall values for the generated highlights.

A relevant area of future research is to evaluate the benefit of the presented text highlighting models through an evaluation with the telehealth users (doctors). 
This type of evaluation can present new insights on how to design word-level text highlighting models for telehealth applications. 
Another area of research is incorporating active learning to the labelling process. 
That is, by using active learning strategies, the dataset preparation time and the labeling process can be improved, and the text highlighting dataset samples can be selected more carefully for labelling.

\section{Acknowledgements}
The authors would like to thank Your Doctors Online for providing funding and support for this research. This work was funded and supported by Mitacs through the Mitacs Accelerate Program.

\bibliographystyle{plainnat} 
\bibliography{Ref}

\end{document}